\documentclass[conf]{new-aiaa}
\usepackage[utf8]{inputenc}

\usepackage{graphicx}
\usepackage{amsmath}
\usepackage[version=4]{mhchem}
\usepackage{siunitx}
\usepackage{longtable,tabularx}
\title{Automatic Parameterization for Aerodynamic Shape Optimization via Deep Geometric Learning}

\author{Zhen Wei~\footnote{Doctoral Student, Computer Vision Lab, School of Computer and Communication Sciences, AIAA Member} }
\author{Pascal Fua~\footnote{Professor, Computer Vision Lab, School of Computer and Communication Sciences}}
\affil{EPFL, Lausanne, 1015, Switzerland}
\author{Michaël Bauerheim~\footnote{Associate Professor, Department of Aerodynamics, Energies and Propulsion}}
\affil{ISAE-SUPAERO, Toulouse, 31055, France}

\newif\ifdraft
\draftfalse
\drafttrue

\usepackage[dvipsnames]{xcolor}
\definecolor{orange}{rgb}{1,0.5,0}
\definecolor{violet}{RGB}{70,0,170}

\ifdraft
 \newcommand{\PF}[1]{{\color{red}{\bf PF: #1}}}
 
 \newcommand{\WZ}[1]{{\color{SeaGreen}{\bf ZW: #1}}}
 \newcommand{\wz}[1]{{\color{SeaGreen} #1}}
 \newcommand{\MB}[1]{{\color{blue}{\bf MB: #1}}}
 
 \newcommand{\BG}[1]{{\color{orange}{\bf BG: #1}}}
 
 \usepackage{ulem}
\else
 \newcommand{\PF}[1]{}
 
 \newcommand{\WZ}[1]{}
 \newcommand{\wz[1]{ #1 }
 \newcommand{\MB}[1]{}
 
 \newcommand{\BG}[1]{}
 
\fi

\newcommand{\comment}[1]{}

\newcommand{\bv}{\mathbf{v}}

\newcommand{\bz}{\mathbf{z}}

\newcommand{\bZ}{\mathbf{Z}}

\newcommand{\cL}{\mathcal{L}}
\newcommand{\cO}{\mathcal{O}}

\newcommand{\argmin}{\operatornamewithlimits{argmin}}

\begin{document}

\maketitle

\begin{abstract}
We propose two deep learning models that fully automate shape parameterization for aerodynamic shape optimization. Both models are optimized to parameterize via deep geometric learning to embed human prior knowledge into learned geometric patterns, eliminating the need for further handcrafting. The Latent Space Model (LSM) learns a low-dimensional latent representation of an object from a dataset of various geometries, while the Direct Mapping Model (DMM) builds parameterization on the fly using only one geometry of interest. We also devise a novel regularization loss that efficiently integrates volumetric mesh deformation into the parameterization model. The models directly manipulate the high-dimensional mesh data by moving vertices. LSM and DMM are fully differentiable, enabling gradient-based, end-to-end pipeline design and plug-and-play deployment of surrogate models or adjoint solvers. We perform shape optimization experiments on 2D airfoils and discuss the applicable scenarios for the two models. Codes will be available at \url{https://github.com/kfxw/CFD_Mesh_Parameterization}.
\end{abstract}

\section{Nomenclature}

{\renewcommand\arraystretch{1.0}
\noindent\begin{longtable*}{@{}l @{\quad=\quad} l@{}}
$C_d$ & the drag coefficient \\
$f_\Theta$ & the Latent Space Model with its weights $\Theta$ \\
$g_\Phi$ & the Direct Mapping Model with its weights $\Phi$ \\
$h$ & the CFD evaluation module \\
$\cL$ & the loss function for the deep geometric training \\
$\cL_{reg}$ & the regularization loss function \\
$\cL_{cons}$ & the geometric constraint used during shape optimization \\
$\hat{M}=\{\hat{V},\hat{E}\}$ & the template mesh and its vertices, edges \\
$M=\{\hat{V}+\Delta V,\hat{E}\}$ & the deformed CFD mesh, generated by $f_\Theta$ or $g_\Phi$ \\
$S$ & the target geometry to be parameterized \\
$\bz$ & the latent vector of the Latent Space Model \\
\end{longtable*}}

\section{Introduction}
\lettrine{G}{eometry} parameterization and manipulation are central to aerodynamic shape optimization problem.
However, effective and efficient parameterization often requires significant human intervention, hindering full automation of shape optimization.
It happens for several reasons.
First, mesh parameterization and hyperparameters depend heavily on the object of interest, requiring geometric prior knowledge and case-by-case analysis.
For example, determining the number and position of control points for the Free Form Deformation (FFD) \cite{Sederberg86, Lamousin94, Kenway10} and Nonuniform Rational B-Splines (NURBS) \cite{Toal10} methods, and the number of polynomial bases for Class/Shape function Transformation (CST)~\cite{Kulfan08} is critical to the performance.
Second, conventional method only parameterize object surfaces, which leads to detachment of the computational mesh from the original surface, necessitating time-consuming remeshing or additional mesh deformation algorithms \cite{deBoer07, Batina90, Batina91, Farhat98, Luke12} that increase computational time and require additional tuning of hyperparameters.
Third, conventional methods are less capable in handling the high dimensionality. 
Dimension reduction methods are often applied to mitigate the curse of dimensionality \cite{Bellman61} but they also introduce new problems.
For example, the use of Proper Orthogonal Decomposition (POD) methods \cite{Robinson01,Poole15, Masters17,Li19c,Kedward20} are usually coupled with task-specific handcraft engineering to reconstruct the geometry from a latent vector.
The Active Subspace methods \cite{Constantine14,Li19,Lukaczyk14,Namura17,Grey18,Bauerheim16,Magri16} exploit the surrogate models' gradients, but the upper bound of its approximation error increases with the dimensionality.
The Generative Topographic Mapping (GTM) method~\cite{Viswanath11} faces an exponential increase in the number of the radius basis to tune manually.
Fourth, none of the existing dimension reduction approaches to automate parameterization \cite{Robinson01, Poole15, Masters17, Li19c, Kedward20, Viswanath11, Constantine14} are designed to be differentiable, making seamless integration into gradient-based frameworks, such as those that involve deep learning or adjoint CFD solvers, impossible.

More recently, several works have emphasized the significance of using deep learning techniques in shape parameterization \cite{li20, Li21,Li22b} and dimension reduction \cite{Glaws22}. However, they are subject to similar limitations as conventional methods, including the need for geometric priors in model design or data processing, non-differentiability with respect to design variables, and the reliance on conventional parameterization methods.

The main contribution of our work is a novel approach to shape parameterization that (i) eliminates the need for handcrafted parameters, (ii) incorporates computational mesh deformation to parameterize the entire CFD mesh, (iii) can directly handle high-dimensional mesh data for maximizing the freedom of geometry manipulation, and (iv) is fully differentiable, enabling straightforward implementation in gradient-based frameworks used for shape optimization or uncertainty quantification studies. We firstly drew inspiration from computer vision research, which has demonstrated the effectiveness of implicit representations \cite{Park19c} that eliminate explicit shape parameterization, and develop models for continuous mesh representation. We propose and investigate two models: the Latent Space Model (LSM) and the Direct Mapping Model (DMM). Both models encode the geometric information of the object of interest as a deformation from a fixed shape template and decode the parameterized design variables into an entire CFD mesh. The two models employ different training strategies and utilize distinct amounts of training data, resulting in unique properties and applicability to different tasks. Furthermore, we devise a novel regularization loss function that acts as an implicit optimizer, guiding both LSM and DMM to properly deform the CFD mesh to preserve the mesh quality for downstream numerical simulation.

The remainder of this paper is structured as follows. Section \ref{sec:method} will provide a detailed description of both models, including their structures, training and inference. Specifically, Section 1.3 will establish the formulation of the novel regularization loss function, and Section \ref{sec:discussion_lsm_dmm} will discuss the applicable scenario of LSM and DMM. In Section \ref{sec:final_exp_res}, we demonstrate the use of LSM and DMM coupled with a differentiable surrogate model and an adjoint solver for 2D airfoil shape optimization.

\section{Method}
\label{sec:method}
\begin{figure}[tb]
    \begin{center}
        \includegraphics[width=1\linewidth]{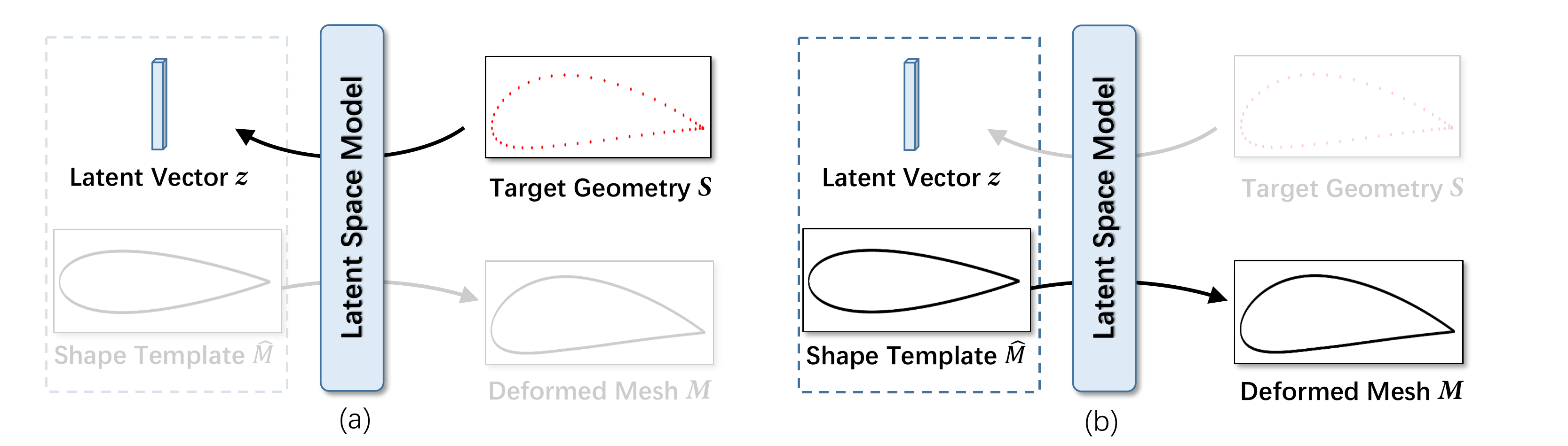}
    \end{center}
    \caption{
        \small The framework of LSM. It first encodes the target geometry into a latent vector, as shown in (a), and then decodes it into a deformation of the shape template, as shown in (b). The latent vector $\bv$ parameterizes the target geometry.
    }
    \label{fig:framework_LSM}
\end{figure}

\begin{figure}[th]
    \begin{center}
        \includegraphics[width=1\linewidth]{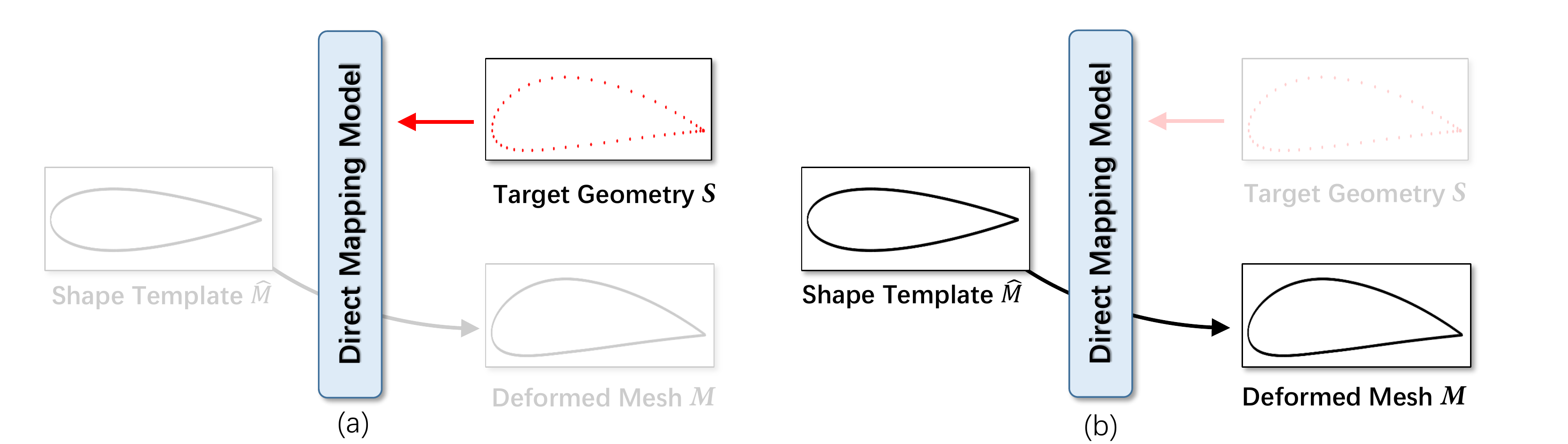}
    \end{center}
    \caption{
        \small The framework of DMM. DMM first learns to fit the target geometry following Eq.\ref{eq:agmin_g_phi}, as shown in (a). Then the model reconstructs the same geometry by deforming the shape template, as shown in (b). The trainable weights $\Phi$ of DMM parameterize the target geometry.
    }
    \label{fig:framework_DMM}
\end{figure}

The proposed models utilize deep geometric learning to offer flexible CFD mesh parameterization.
They are designed to encode a target shape and then deform a given CFD mesh built on a fixed shape template to reconstruct the target geometry. Only points sampled on the target's surface are fed into the pipeline, which then generates an entire CFD mesh by inferring the model.
The pipelines of LSM and DMM are shown in Fig.\ref{fig:framework_LSM} and Fig.\ref{fig:framework_DMM}, respectively.

To formalize this process, we consider a 2D template airfoil with two sets of sampled points: $\hat{V}^S = \{\hat{\bv}^S_1, \hat{\bv}^S_2,..., \hat{\bv}^S_n$\} on the surface and $\hat{V}^V = \{\hat{\bv}^V_1, \hat{\bv}^V_2,..., \hat{\bv}^V_m$\} from its surrounding CFD computational domain, respectively. The sampling sizes can be arbitrary. The same vertices in a deformed continuous coordinate space can be represented as $V=\hat{V}+\Delta V$, where $\Delta V = \{\delta {\bv}_1, \delta {\bv}_2,...\}$ is the displacement vector of vertices.
During training, we compute the difference between the $O$ points sampled from the surface of the target airfoil to be parameterized, as denoted by $S=\{\textbf{s}_1, \textbf{s}_2,..., \textbf{s}_o\}$, and $\hat{V}^S$ as the supervised loss, while using $\hat{V}^V$ for unsupervised optimization in the regularization loss.
At the inference stage, $\hat{V}^S$ and $\hat{V}^V$ are resampled according to a user-defined CFD mesh $\hat{M}={\{\hat{V}^S, \hat{V}^V}, \hat{E}\}$ where $\hat{E}$ defines the edges connecting all vertices. Both LSM and DMM perform mesh deformation and infer a deformed CFD mesh $M={\{\hat{V}^S + \Delta V^S, \hat{V}^V + \Delta V^V}, \hat{E}\}$ with the mesh topology fixed.

\subsection{The Latent Space Model (LSM)}
LSM relies on the auto-decoder structure \cite{Tan95, Park19c} and is trained jointly with the latent space on the collected geometries that belong to a same category (e.g. 2D airfoils). It maps a CFD mesh built on the template airfoil conditioned on a latent vector that parameterizes the target geometric information as
\begin{align}
    f_{\Theta} : \mathbb{R}^d \times \mathbb{R}^D \rightarrow \mathbb{R}^D \; , \label{eq:map_lsm} \\
    \mathbf{\delta v} = f_{\Theta}(\bz, \hat{\bv}) \; , \nonumber
\end{align} 
where $D$ is the data dimension of CFD mesh, $d \ll D$, and $\bz\in \mathbb{R}^d$ is the low-dimensional latent parameterization. $\Theta$ represents the weights that control LSM $f$. 

During training, a training dataset of $T$ airfoils $S_1,\ldots,S_T$ is collected which only contains sampled surface points. The auto-decoding approach~\cite{Park19c, Tan95} is used to optimize the weights $\Theta$ and the latent vectors that parameterize each training data $\textbf{z}_1,\ldots,\textbf{z}_T$. The training objective writes
\begin{align}
\Theta^*,\bZ^* &=  \argmin_{\Theta,\bz_1,\ldots,\bz_T} \sum_{t=1}^T \cL(f_{\Theta}(\bz_t, \hat{\bv}^S), f_{\Theta}(\bz_t, \hat{\bv}^V), S_t) \; ,
\label{eq:argmin_f_z_theta}
\end{align}
where $\cL$ is a loss function that is small when $f_{\Theta}(\bz_t, \hat{\bv}^S)$ yields a deformed airfoil that is geometrically identical to $S_t$, and $f_{\Theta}(\bz_t, \hat{\bv}^V)$ generates a CFD mesh $M$ that possess satisfactory quality. As a result, the optimal $\bz_t^*$ corresponds to a low-dimensional parameterization of $S_t$. 

We use the Chamfer Distance \cite{Barrow77}, denoted as $\cL_{CD}(V^S, S)$, to measure the geometric difference and the regularization loss, denoted as $\cL_{reg}(V^V)$, to preserve the quality of the computational mesh. Additionally, we apply a constraint on the norm of $\bz$ to ensure the smoothness of the learned latent space.
Therefore, we have the overall training objective that writes
\begin{align}
\cL(V^S, V^V, S) = \cL_{CD}(V^S, S) + w_{reg}\cL_{reg}(V^V) + w_{\bz}\left|\left|\bz\right|\right|^2 ,
\end{align}
where $w_{reg}$ and $w_{\bz}$ are the balancing weights.

At inference time, given a target geometry $S$ and the frozen weights $\Theta^*$, the latent parameterization vector $\bz^*$ is obtained by optimizing the following problem,
\begin{align}
\bz^* &=  \argmin_{\bz} \cL(f_{\Theta}(\bz, \hat{\bv}^S), f_{\Theta}(\bz, \hat{\bv}^V), S) \; .
\label{eq:agmin_f_z}
\end{align}
Then the output CFD mesh is obtained by forwarding $\bz^*$ again as $M=\{\{f_{\Theta}(\bz^*, \hat{\bv}^S), f_{\Theta}(\bz^*, \hat{\bv}^T)\}, \hat{E}\}$. 

\subsection{The Direct Mapping Model (DMM)}
DMM is a feed-forward neural network dedicated to map the template airfoil to a single target, which writes
\begin{align}
    g_{\Phi} : \mathbb{R}^D \rightarrow \mathbb{R}^D \; , \label{eq:map_dmm} \\
    \mathbf{\delta v} = g_{\Phi}(\hat{\bv}) \; , \nonumber
\end{align} 

Unlike LSM, which uses an auto-decoder model and $T$ exclusive latent vectors to parameterize $T$ shapes of the same category, DMM requires $T$ neural networks $g_{\Phi_1}, g_{\Phi_2},\ldots, g_{\Phi_T}$ for $T$ template-to-target geometry mappings.
In DMM, the geometric information is embedded in the model itself and the model's weights parameterize the target geometry without producing an explicit vector as design variables. However, DMM does not have a training process or a requirement for a large amount of data. During inference, $\Phi$ is randomly initialized and then optimized with respect to the target airfoil $S$ to solve the following problem.
\begin{align}
\Phi^* &=  \argmin_{\Phi} \cL(g_{\Phi}(\hat{\bv}^S), g_{\Phi}(\hat{\bv}^V), S) \; .
\label{eq:agmin_g_phi}
\end{align}

DMM's loss function is similar to LSM's loss function, but without the constraint on the latent space, which writes
\begin{align}
\cL(V^S, V^V, S) = \cL_{CD}(V^S, S) + w_{reg}\cL_{reg}(V^V).
\end{align}

\subsection{The Regularization for Quality-Preserved Mesh Deformation}
The regularization loss guarantees that all points sampled as $\hat{V}^V$ move in accordance with the changing airfoil geometry to maintain the computational mesh quality inherent in the user-defined CFD mesh that is built on the template airfoil. This quality is reflected in quantitative metrics, such as cells' skewness, orthogonality, and aspect ratio, as well as the mesh's ability to represent the underlying physics~\cite{knupp07}.
To this end, we characterize the computational mesh as a discrete sample of a continuous 2D elastic material which is framed by the fixed computational boundaries (typically the inlet/outlet patches) and the deformable airfoil surface.
Mesh quality can degrade due to non-rigid distortion, which changes the angles and edge length ratios within the mesh cells. 
Deformation of computational boundaries can lead to incorrect boundary conditions and hinder the use of generated meshes in downstream software.
To address these issues, we formulate the regularization loss as the sum of two terms, namely $\cL_{dist}$ and $\cL_{def}$, to minimize non-rigid distortion across the entire mesh and the deformation of fixed patches:
\begin{equation}
    \cL_{reg} = \cL_{dist} + \cL_{def}\;\;.
\end{equation}

To formulate $\cL_{dist}$, we define an energy function $\mathbb{E}$ that quantifies the total non-rigid distortion as the squared Frobenius norm of the infinitesimal strain tensor, which writes
\begin{align}
    \begin{split}
        \mathbb{E} &:= \frac{1}{2} \sum_{i=1}^{|V^V|} \left|\left| \nabla \delta \bv^V_i + (\nabla \delta \bv^V_i)^T \right|\right|_F^2  \\
        &\;= \frac{1}{2} \sum_{i=1}^{|V^V|} \left\{ 
        \left|\left| \frac{\partial \delta v^V_{x,i}}{\partial x} \right|\right|^2 + 
        \left|\left| \frac{\partial \delta v^V_{x,i}}{\partial y} \right|\right|^2 + 
        \left|\left| \frac{\partial \delta v^V_{y,i}}{\partial x} \right|\right|^2 + 
        \left|\left| \frac{\partial \delta v^V_{y,i}}{\partial y} \right|\right|^2 \right\} \;\;,
    \end{split}
\end{align}
where the displacement of a single vertex $\delta \bv^V_i = (\delta v^V_{x,i}, \delta v^V_{y,i}) \in \mathbb{R}^2$. Then we investigate the following formula
\begin{equation}
    - \sum_{i=1}^{|V^V|} \left\{
    \frac{\partial^2 \delta v^V_{x,i}}{\partial x^2} +
    \frac{\partial^2 \delta v^V_{x,i}}{\partial y^2} +
    \frac{\partial^2 \delta v^V_{y,i}}{\partial x^2} +
    \frac{\partial^2 \delta v^V_{y,i}}{\partial y^2}
    \right\} \;\;.
    \label{eq:avm_internal_step}
\end{equation}
Since $\delta \bv_i = \bv_i - \hat{\bv}_i$ and $\partial ^2 \hat{\bv}_i / \partial x^2 = \partial ^2 \hat{\bv}_i / \partial y^2 = 0$, Eq.\ref{eq:avm_internal_step} can be rewritten as
\begin{equation}
    F := - \sum_{i=1}^{|V^V|} \left\{
    \frac{\partial^2 v^V_{x,i}}{\partial x^2} +
    \frac{\partial^2 v^V_{x,i}}{\partial y^2} +
    \frac{\partial^2 v^V_{y,i}}{\partial x^2} +
    \frac{\partial^2 v^V_{y,i}}{\partial y^2}
    \right\} \;\;.
\end{equation}
$F=0$ for $\forall i$ when $\bv = \hat{\bv}$, which is the Euler-Lagrange equation of $\mathbb{E}$ and is the sufficient and necessary condition of which the user-defined template CFD mesh is a stationary point on $\mathbb{E}$'s energy landscape. $\cL_{dist}$ is then defined to induce local cell structures in the generated computational mesh to resemble those in the template, so as to prevent the creation of meshes with serious issues such as negative-volume cells and severely non-orthogonal issues. The desired mesh properties embedded in the template mesh, including cell aspect ratio, skewness, and orthogonality, are also preserved as much as possible.
$\cL_{dist}$ is defined as
\begin{equation}
    \cL_{dist} := \sum_{i=1}^{|V^V|} \left\{
    \left|\left| \frac{\partial^2 v^V_{x,i}}{\partial x^2} \right|\right|^2 + 
    \left|\left| \frac{\partial^2 v^V_{x,i}}{\partial y^2} \right|\right|^2 + 
    \left|\left| \frac{\partial^2 v^V_{y,i}}{\partial x^2} \right|\right|^2 + 
    \left|\left| \frac{\partial^2 v^V_{y,i}}{\partial y^2} \right|\right|^2
    \right\} \;\;.
\end{equation}
Thanks to the full differentiation of LSM and DMM, the PDE terms can be easily computed analytically through auto-differentiation.

$\cL_{def}$ measures the deformation near the boundaries, which writes
\begin{equation}
    \cL_{def} := \sum_{j=1}^{|V^{fix}|} \left|\left| \delta \bv^{fix} \right|\right|^2 \;\;,
\end{equation}
where $V^{fix}=\{\bv^{fix}_1,\bv^{fix}_2,\ldots,\bv^{fix}_k\}$ is a set of sampled points on or closed to the computational boundaries (i.e. inlet/outlet patches) and any other specified fixed patches, and is a subset of $V^V$.

\subsection{Discussions on LSM and DMM}
\label{sec:discussion_lsm_dmm}
\subsubsection{The Commonness of LSM and DMM}
\begin{figure}[!htb]
    \begin{center}
        \includegraphics[width=1\linewidth]{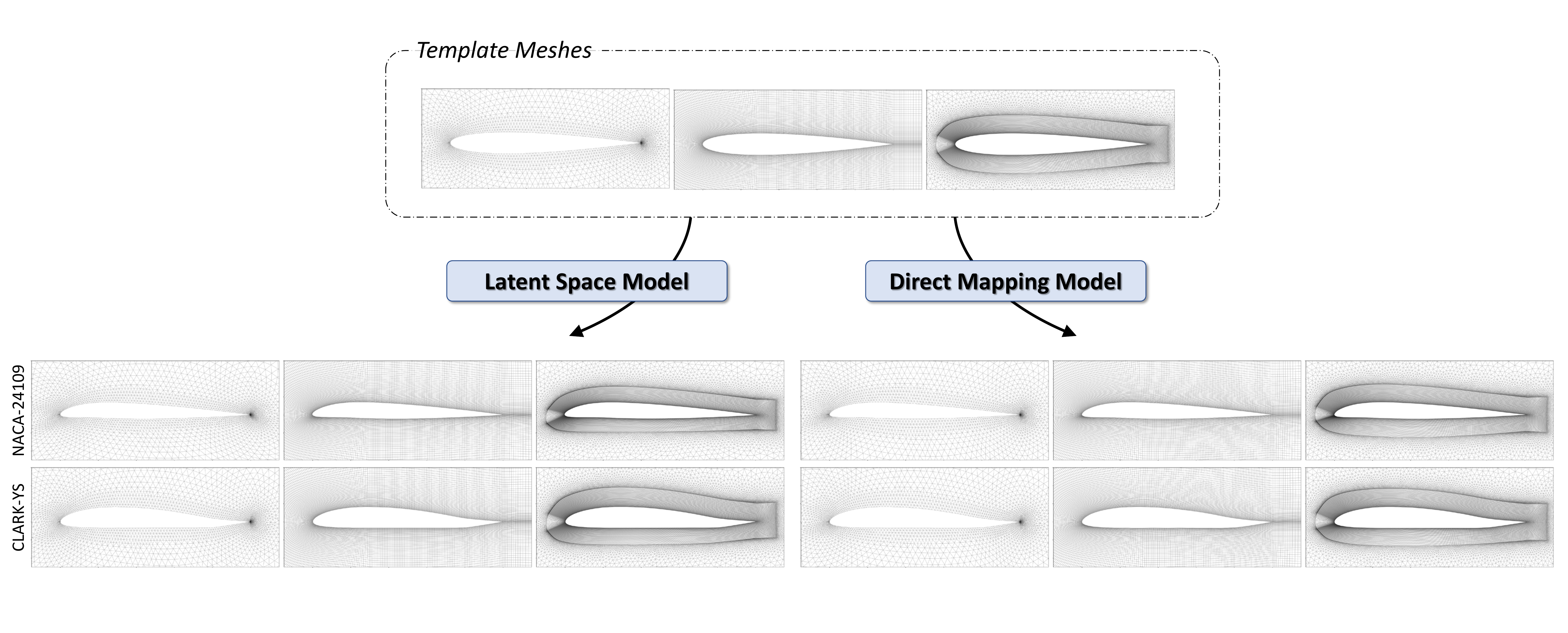}
    \end{center}
    \caption{
        \small Both optimized LSM and DMM can infer various template CFD meshes accurately without requiring any adaptations.
    }
    \label{fig:infer_different_templates}
\end{figure}

LSM and DMM learn continuous mappings between coordinate spaces. The sampling strategy of $V^S$ and $V^V$ can be independent of any specific CFD mesh, or can use a CFD mesh to sample but to generalize to any other discrete sampling in the continuous space.
This enables both models to infer with various CFD meshes after optimizing $\Theta$ or $\Phi$ without any adaptation.
Fig.\ref{fig:infer_different_templates} demonstrates the reconstruction results of both models inferred with different template meshes.

Meanwhile, $\cL_{reg}$ incorporates the mesh deformation into the parameterization models, eliminating the need for an additional postprocessing module. It does not introduce sensitive hyperparameters to tune or manual configurations. Additionally, compared to RBF-based methods and spring analog methods, $\cL_{reg}$ does not need to solve a huge linear system of vertices. It is also more computationally efficient than IDW-based methods. Let $N^*$ and $M^*$ be the numbers of vertices of the geometry surface and the computational mesh for a specific CFD mesh, respectively. The number of sampled $V^V$ is $M$, and $M<<M^*$. The complexity of $\cL_{reg}$ is $\cO(M)$ during training and is none during inference, whereas it is $\cO(N^*M^*)$ for IDW-based methods for every deformation.

\begin{figure}[!tb]
    \begin{center}
        \includegraphics[width=0.75\linewidth]{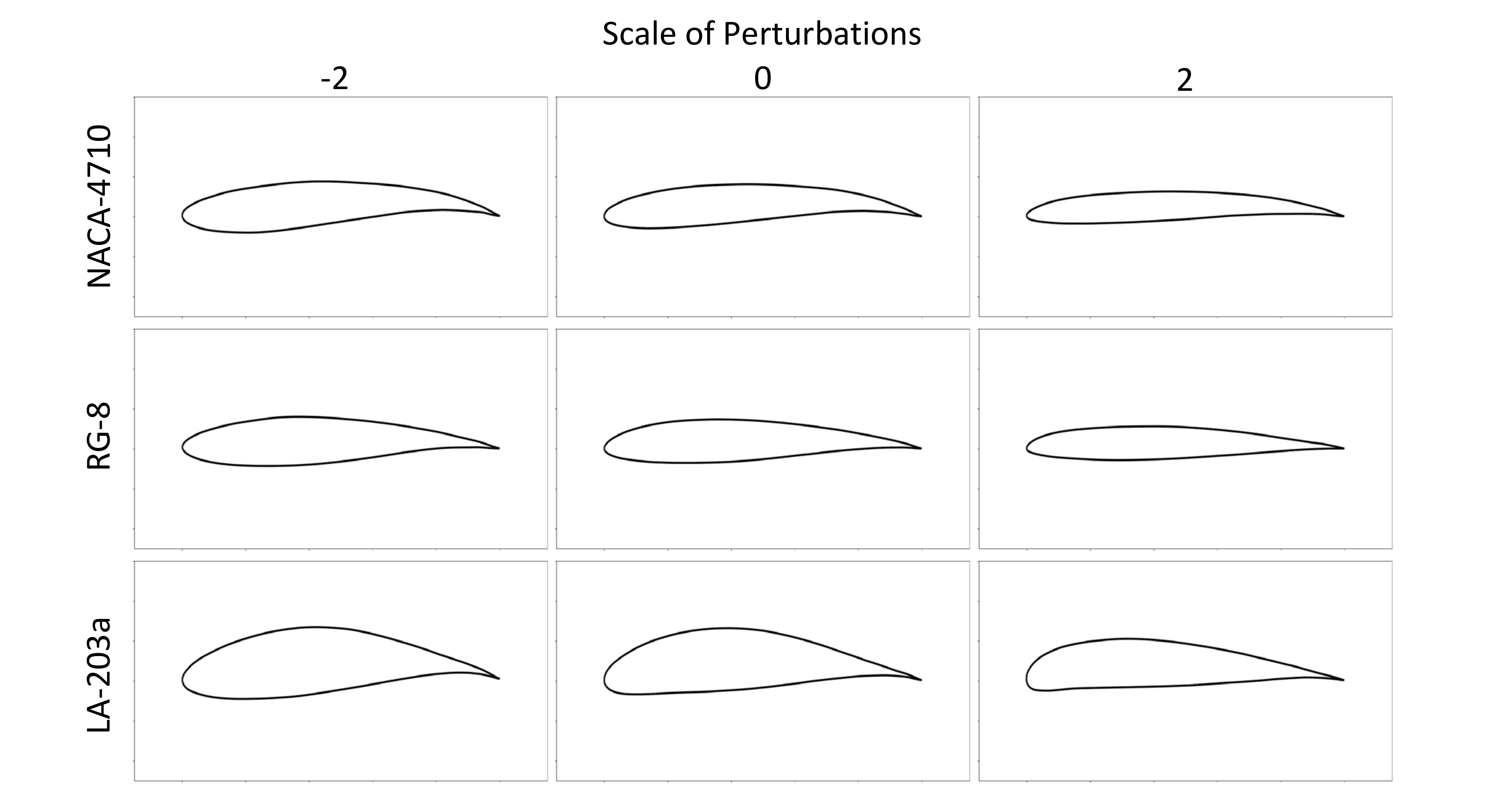}
    \end{center}
    \caption{
        \small The novel airfoils sampled in LSM's learned latent space by perturbing the latent parameterization vectors of existing profiles with the sampling method proposed by \textit{Shen et al.} \cite{shen21}.
    }
    \label{fig:lsm_sample}
\end{figure}
\begin{figure}[!tb]
    \begin{center}
        \includegraphics[width=1\linewidth]{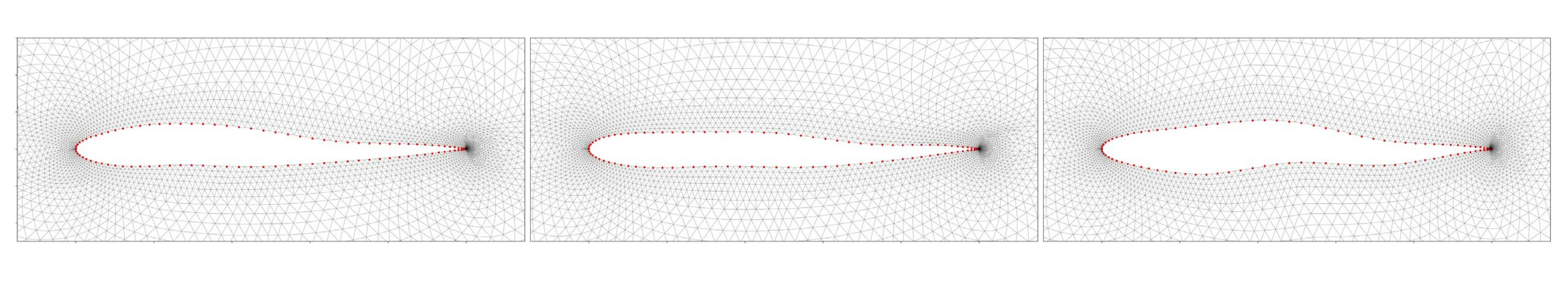}
    \end{center}
    \caption{
        \small The reconstruction results of DMM that parameterize the randomly deformed airfoils. The true geometries of the targets are represented by the red dots.
    }
    \label{fig:dmm_random_reconstruct}
\end{figure}

\subsubsection{The Unique Properties of LSM}
LSM and DMM have different properties due to their different optimization strategies.

The latent space jointly trained with LSM's network embeds rich geometric information from the training data. Novel shapes generated by LSM interpolate the design space defined by the training set and inherit the geometric prior. When perturbing $\bz$ of an existing airfoil slightly, one can obtain another novel airfoil instead of any non-airfoil shape. This property has two major benefits. First, LSM can be used as a novel data sampler. Fig.\ref{fig:lsm_sample} demonstrates the explored novel shapes from existing airfoils by LSM using the sampling-free approach proposed in \cite{shen21}. Second, when performing shape optimization with LSM, we can use fewer explicit geometric constraints to ensure the validity of the optimized result, such as maintaining the surface smoothness, object scale and the special structures commonly appeared among the training data.

LSM is an upgraded version of the model proposed in Wei et al. \cite{wei23} with a simplified structure and regularization loss, resulting in less computational time and memory consumption during both training and inference.

\subsubsection{The Unique Properties of DMM}
DMM is a lightweight model suitable for a case study of a specific geometry of interest. It does not require a training dataset, which is not always available. Optimizing DMM takes $4-6s$ for 2D airfoil cases. Meanwhile, DMM encodes the geometric information of a single object, which allows for more geometric freedom and the ability to deform the template shape into various geometries. Fig.\ref{fig:dmm_random_reconstruct} demonstrates that DMM can reconstruct and parameterize randomly twisted shapes from the template airfoil NACA-0012. Using DMM as the parameterization model, shape optimization can explore and develop novel structures.

\subsection{Implementation}
The LSM and DMM are implemented using the Pytorch deep learning framework \cite{paszke19}. LSM is a four-layer multi-layer perceptron (MLP) network with a $256$-dimensional latent space. The activation function of each latent MLP layer is a weighted sum of ReLU \cite{fukushima75} and SIREN \cite{sitzmann20}, which enables accurate shape deformation and the computation of second-order partial derivatives. LSM is trained on $2648$ airfoils, including $1442$ data of various foil series collected from the UIUC airfoil dataset \cite{Selig96a} and the remaining airfoils being random NACA 4-digit or 5-digit profiles. The training uses the Adam optimizer \cite{Kingma15} with an initial learning rate of $10^{-4}$ for 20 epochs, which costs $175s$.
During inference, the latent vector $\bz$ is optimized with the Adam optimizer with at maximum $1000$ iterations.

DMM uses a similar network structure as LSM, but with only two latent layers. During inference, the Adam optimizer solves Eq.\ref{eq:agmin_g_phi} with $600$ iterations.

\begin{figure}[tbh]
    \begin{center}
        \includegraphics[width=1\linewidth]{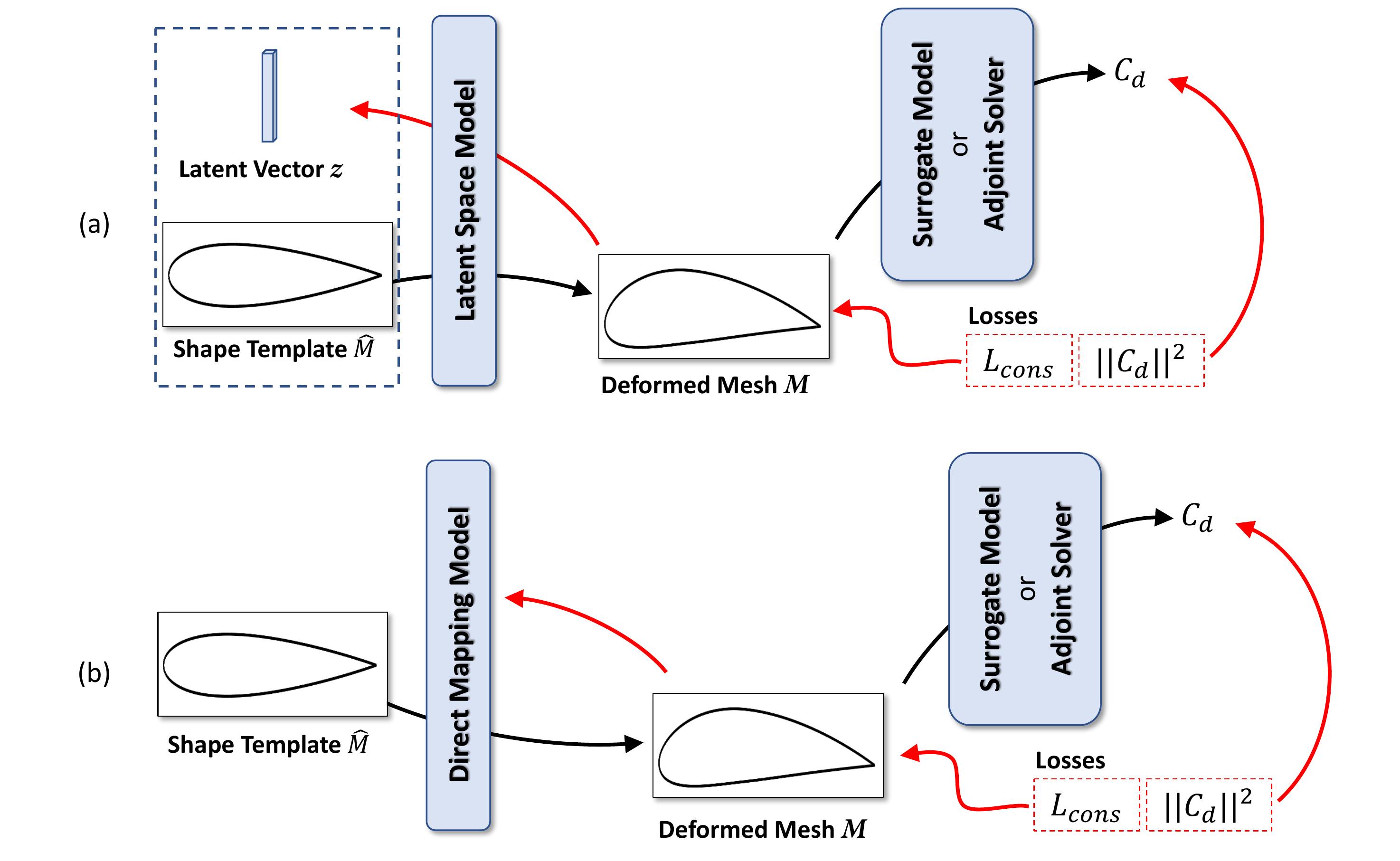}
    \end{center}
    \caption{
        \small The shape optimization frameworks with the use of (a) LSM and (b) DMM. The surrogate model or the adjoint solver is appended after the inferred meshes. Red paths illustrate the gradient flows in the pipeline.
    }
    \label{fig:framework_optim}
\end{figure}

\subsection{Fully Differentiable Frameworks for Shape Optimization}
In this paper, we use LSM and DMM parameterization models to provide an example of solving the ADODG Case 1 optimization problem with a gradient-based solution. The objective is to minimize the drag coefficient of the NACA-0012 airfoil in a transonic flow (Ma=0.85) at zero angle-of-attack.

To perform aerodynamic shape optimization, we need to couple LSM or DMM with a downstream evaluation module (denoted as $h$), which can be either a differentiable surrogate model (such as the GCNN network~\cite{Baque18}) or an adjoint solver (such as SU2~\cite{Economon16}), to evaluate the performance of the current design. For example, the evaluate of drag can be writted as
\begin{equation}
    C_d = h(M)\;,
\end{equation}
where $M=\{\{\hat{V}^S+\Delta V^S, \hat{V}^V+\Delta V^V\}, \hat{E}\}$ is inferred by $f_{\Theta}$ or $g_{\Phi}$.
The geometric constraints $\cL_{cons}$ can be directly calculated from $M$ and then applied to $M$. The overall pipelines are illustrate in Fig.\ref{fig:framework_optim}. The shape optimization objectives are written as
\begin{align}
    \text{for LSM: }&\;\;\; \bz^* =   \argmin_{\bz} \left( \left|\left|C_d\right|\right|^2 + \cL_{cons}(M(\hat{V} + f_{\Theta}(\hat{M},\bz))) \right)\; ,\\
    \text{for DMM: }&\;\;\;  \Phi^* =   \argmin_{\Phi} \left( \left|\left|C_d\right|\right|^2 + \cL_{cons}(M(\hat{V} + g_{\Phi}(\hat{M},\bz))) \right)\; .
\end{align}

The gradients from the surrogate model or the adjoint solver, as well as the constraints, will be backpropagated to the latent code of LSM or to the weights of DMM. The output shapes are then manipulated by applying an optimizer to update the learnable parameters.
\section{Experiments and Results}
\label{sec:final_exp_res}
We investigate the effectiveness of the proposed models on 2D airfoil optimization cases following the settings in ADODG Case 1. The optimization objective is to reduce the drag coefficient of NACA-0012 in a transonic inviscid flow (Ma=0.85) at zero angle-of-attack. To demonstrate the benefits of the proposed model's full differentiability, we use different methods to evaluate the airfoil's drag coefficient and generate the CFD's gradients (i.e. surface sensitivities or uncertainties) with respect to the geometry. Specifically, we use a GCNN \cite{Baque18} surrogate model and SU2's adjoint solver \cite{Economon16}, respectively.
In the remainder of this paper, we use terms like \textit{LSM + SU2} to represent a pair of parameterization model and evaluation model.

\subsection{The Evaluation Modules}
\subsubsection{The GCNN Surrogate Model}
The GCNN takes the airfoil's connected contour as input and extracts geometric features based on the vertices and edges. It predicts the drag coefficient in two ways. ${C_d}^{int}$ is obtained by predicting air pressure values per vertex and computing an integral of pressure over the airfoil's surface and ${C_d}^{direct}$ is obtained by predicting an overall drag coefficient directly. The minimization objective is a summation of both predictions as $C_d = {C_d}^{int} + {C_d}^{direct}$.

The GCNN network structure consists of 5 graph convolution blocks, each with 3 convolutional layers. Batch normalization \cite{Sergey15} is used after each layer and the Exponential Linear Unit \cite{clevert16} is used as an activation function. The model is trained using the Adam optimizer with a learning rate of $5 \times 10^{-4}$ for 900 epochs.

\subsubsection{The SU2 Adjoint Solver}
We use SU2's continuous adjoint solver to calculate the sensitivity scale of the drag minimization objective along the surface's vertex normal direction using the simulation result. To avoid the non-uniqueness issue of the Euler equation \cite{Masters17,LeDoux15}, we employ a restart strategy for flow field initialization \cite{he19}. During optimization, we gradually change the airfoil's geometry between successive iterations with a moderate learning rate for the parameterization model to ensure the optimization result converges to a satisfying minimum. We speedup this process by updating the adjoint result at regular intervals and reusing the SU2 gradient for multiple iterations, given that adjoint solving takes considerably more time than evaluating a surrogate model. In case of numerical instability, we discard the SU2's gradient if its norm becomes excessively large, even if the minimum residual criteria are met.

\subsection{The Geometric Constraints}
The overall geometric constraint can be generically expressed as a weighted sum of $P$ task specific constraints, $Q$ parameterization model specific constraints and $R$ evaluation model specific constraints: $\cL_{cons} = \sum_{i=1}^P w_{T,i}C_{T,i} + \sum_{j=1}^Q w_{M,j}C_{M,j} + \sum_{k=1}^R w_{E,k}C_{E,k}$, where $C_T$ is defined by the optimization objective, $C_M$ is defined by the parameterization models and $C_E$ is defined by the evaluation modules. $w_{T,i}$, $w_{M,j}$ and $w_{E,k}$ are weights for the corresponding constraints.

\subsubsection{Task Specific Constraints}
We denote the geometric constraint of ADODG Case 1 as the bounding constraint $C_{T,1}$, which limits the airfoil’s thickness from decreasing below its initial value and writes
\begin{equation}
    {C_{T,1}} = \frac{1}{{|{V^S}|}}\sum\limits_{i = 1}^{|{V^S}|} \left|\left|{\max (|{v}_{y,i}^{S,opt}| - |{v}_{y,i}^{S,init}|,0)}\right|\right|^2\;\;,
\end{equation}
where ${v}_{y,i}^{S,opt}$ and ${v}_{y,i}^{S,init}$ are $y$ coordinates of surface vertices from the optimized airfoil and the initial airfoil, respectively.

\begin{figure}[!tb]
    \begin{center}
        \includegraphics[width=0.9\linewidth]{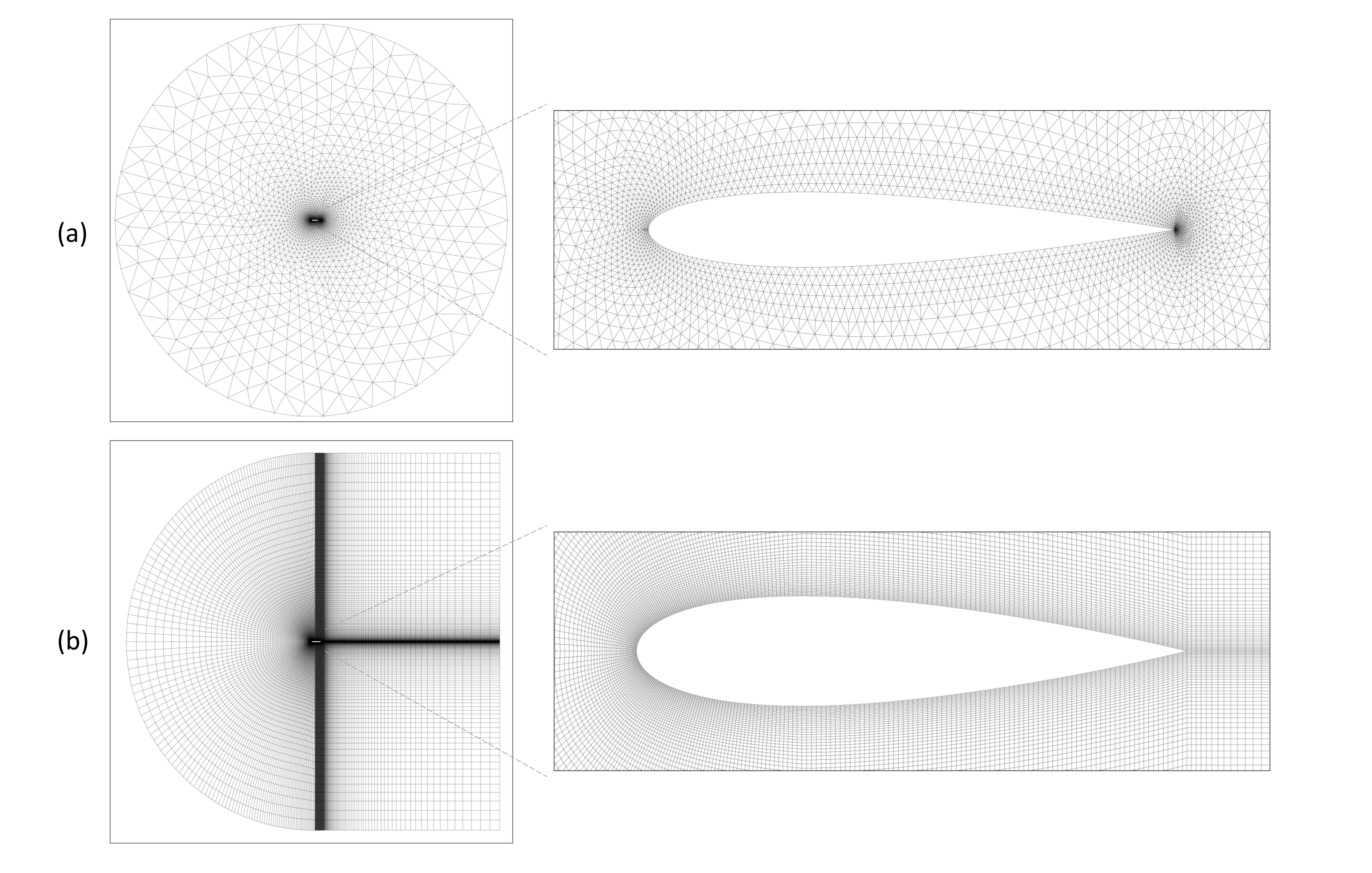}
    \end{center}
    \caption{
        \small The two template meshes employed in the adjoint solver based shape optimization experiments: (a) \textit{TM-A} and (b) \textit{TM-B}. Both meshes are built on NACA-0012.
    }
    \label{fig:template_meshes}
\end{figure}

\subsubsection{Parameterization Model Specific Constraints}
For LSM, we define a constraint on the norm of latent code $\bz$, i.e. $C_{M,1}$, to ensure that the optimized $\bz^*$ falls within the latent distribution learned from the training data, and the generated airfoil is valid, which writes
\begin{equation}
    C_{M,1} = ||\bz||^2_F\;\;.
\end{equation}

For DMM, we fix the leading edge $\bv_{leanding}^S$ and trailing edge $\bv_{trailing}^S$ and constrain the $x$ coordinate to maintain the normalized chord length and chord position. We denote these constraints as $C_{M,2}$ and $C_{M,3}$, respectively, which write
\begin{equation}
    C_{M,2}=\left|\left| \bv_{leanding}^S - [0,0] \right|\right|^2 + \left|\left| \bv_{trailing}^S - [1,0] \right|\right|^2\;\;,
\end{equation}
\begin{equation}
    C_{M,3}=\sum_i^{|V^S|} \left|\left| \max (0 - \bv_{x,i}^S, 0) + \max (\bv_{x,i}^S - 1, 0) \right|\right|^2\;\;.
\end{equation}

\subsubsection{Evaluation Module Specific Constraints}
Since SU2 calculates the sensitivity scale along the surface normal direction, applying this gradient to the mesh model will result in a redistribution of surface vertices, leading to subsequent non-orthogonality in the volumetric cells directly connected to them. To address this issue, we define a constraint $C_{E,1}$ that decays the horizontal shifting as
\begin{equation}
    C_{E,1} = \sum_i^{|V^S|} \left|\left| \delta\bv_{x,i}^S
 \right|\right|^2\;\;.
\end{equation}
The GCNN does not require any additional constraints.

\subsubsection{Constraints on Computational Mesh}
When both models are combined with an adjoint solver, we again use the regularization loss as a constraint on the quality of the computational mesh, specifically $C_{M,4}=\cL_{reg}$. 

In summary, \textit{LSM + GCNN} uses $\{C_{T,1},C_{M,1}\}$, \textit{LSM + SU2} uses $\{C_{T,1},C_{M,1},C_{E,1}\}$, \textit{DMM + GCNN} uses $\{C_{T,1},C_{M,2},C_{M,3}\}$ and \textit{DMM + SU2} uses $\{C_{T,1},C_{M,2},C_{M,3},C_{E,1}\}$.

\begin{table}[tbp]
  \centering
  \caption{Quantitative comparison with previous shape optimization methods with handcrafted and semi-automatic parameterizaations on ADODG Case I benchmark. The drag coefficients are reported in counts. (1 count = $10^{-4}$)}
    \begin{tabular}{l|c|c}
    \hline
    Models & Initial $C_d$ & Optimized $C_d$ \\
    \hline
    He et al. \cite{he19} & 470.36  & 7.60  \\
    He et al. \cite{he19} & 470.36  & 15.54  \\
    He et al. \cite{he19} & 470.36  & 23.53  \\
    Masters et al. \cite{Masters16} & 469.60  & 25.00  \\
    Bisson and Nadarajah \cite{Bisson15} & 464.20  & 26.20  \\
    He et al. \cite{he19} & 470.36  & 34.33  \\
    Carrier et al. \cite{Carrier14} & 471.20  & 36.70  \\
    Lee etal. \cite{Lee15a} & 457.33  & 42.24  \\
    He et al. \cite{he19} & 470.36  & 50.49  \\
    \textit{DMM + SU2} & 466.11  & 69.32  \\
    Zhang et al. \cite{Zhang16} & 481.28  & 73.08  \\
    Poole et al. \cite{Poole15a} & 469.42  & 83.80  \\
    LeDoux et al. \cite{LeDoux15} & 471.30  & 84.50  \\
    Gariepy et al. \cite{Gariepy15} & 481.60  & 141.70  \\
    \textit{DMM + GCNN} & 466.11  & 278.23  \\
    \textit{LSM + SU2} & 466.11  & 283.93  \\
    Fabiano and Mavriplis \cite{Fabiano16} & 466.96  & 297.02  \\
    \textit{LSM + GCNN} & 466.11  & 354.91\\
    \hline
    \end{tabular}%
  \label{tab:result}%
\end{table}%
\begin{figure}[!htb]
    \begin{center}
        \includegraphics[width=1\linewidth]{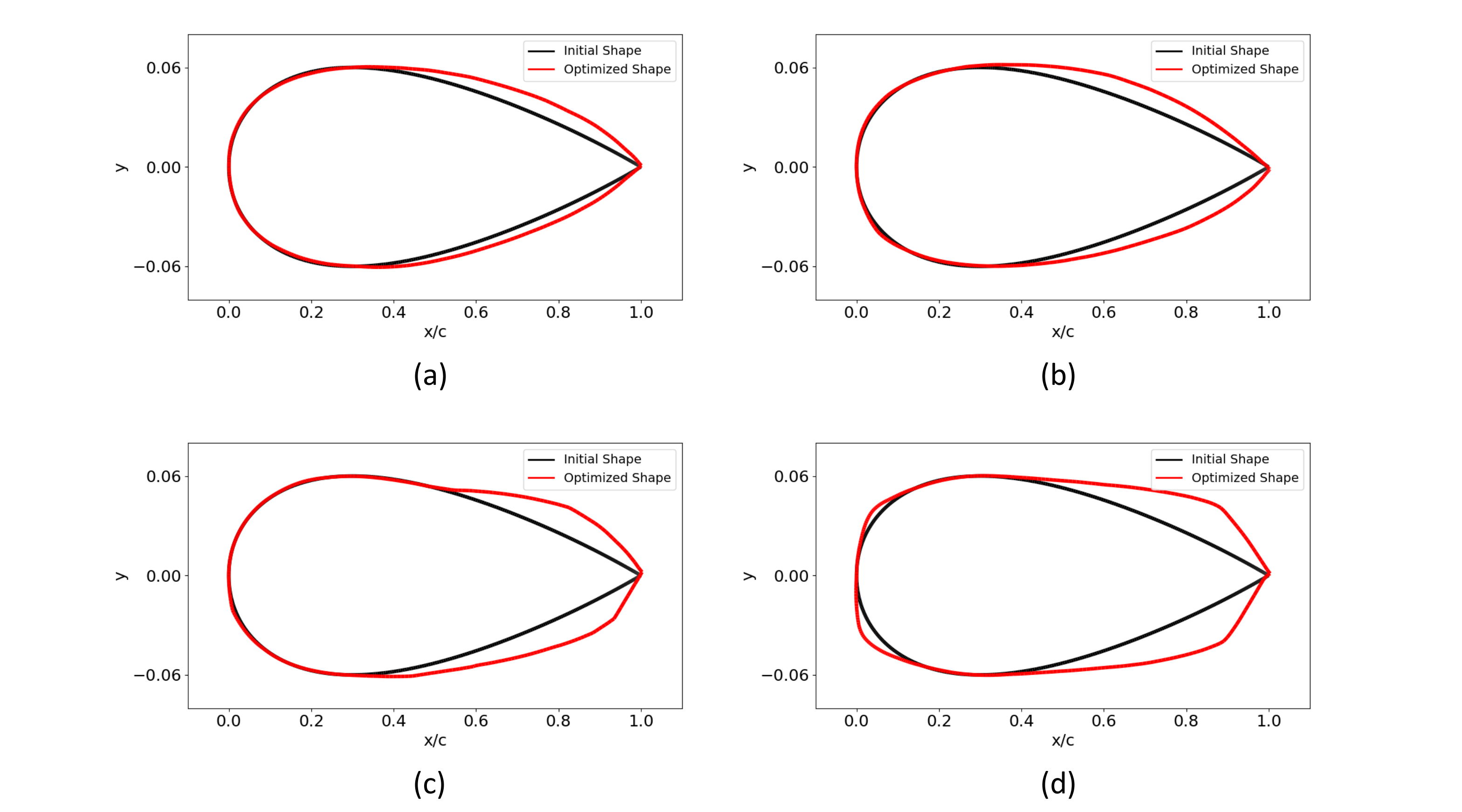}
    \end{center}
    \caption{
        \small The shape optimization results of (a) \textit{LSM + GCNN}, (b) \textit{LSM + SU2}, (c) \textit{DMM + GCNN} and (d) \textit{DMM + SU2}. The initial NACA-0012 and optimized shapes are colored in black and red.
    }
    \label{fig:final_optim_results}
\end{figure}

\subsection{Shape Optimization Results}
When using SU2 for optimization, we accelerate the process by using the coarse template mesh \textit{TM-A}~\footnote{\url{https://github.com/su2code/Tutorials/blob/master/design/Inviscid_2D_Unconstrained_NACA0012/mesh_NACA0012_inv.su2}} for most iterations. Once the drag coefficient stops reducing, we switch to a fine template mesh \textit{TM-B}~\footnote{\url{http://www.wolfdynamics.com/tutorials.html?id=148}} for the final iterations. Fig.\ref{fig:template_meshes} illustrates these template meshes. \textit{TM-A} consists of $5233$ vertices and $10126$ triangles, while \textit{TM-B} consists of $58120$ vertices and $57600$ quadrilateral cells.
When using GCNN for optimization, the choice of template mesh is not important, as GCNN only requires the airfoil's contour as input.

Meanwhile, we introduce a reparameterization mechanism to prevent catastrophic mesh errors terminating the iterations during optimization. When $C_{M,4}$ becomes large or both simulation and adjoint solving fail to meet the residual criteria, we sample points $S'$ from the current airfoil surface $f_{\Theta}(\bz,\hat{\bv}^S)$ or $g_{\Phi}(\hat{\bv}^S)$ and update $\bz$ for LSM or $\Phi$ for DMM using Eq. \ref{eq:agmin_f_z} or Eq. \ref{eq:agmin_g_phi}. Then the optimization continues on the reparameterized model.

\begin{figure}[!htb]
    \begin{center}
        \includegraphics[width=0.9\linewidth]{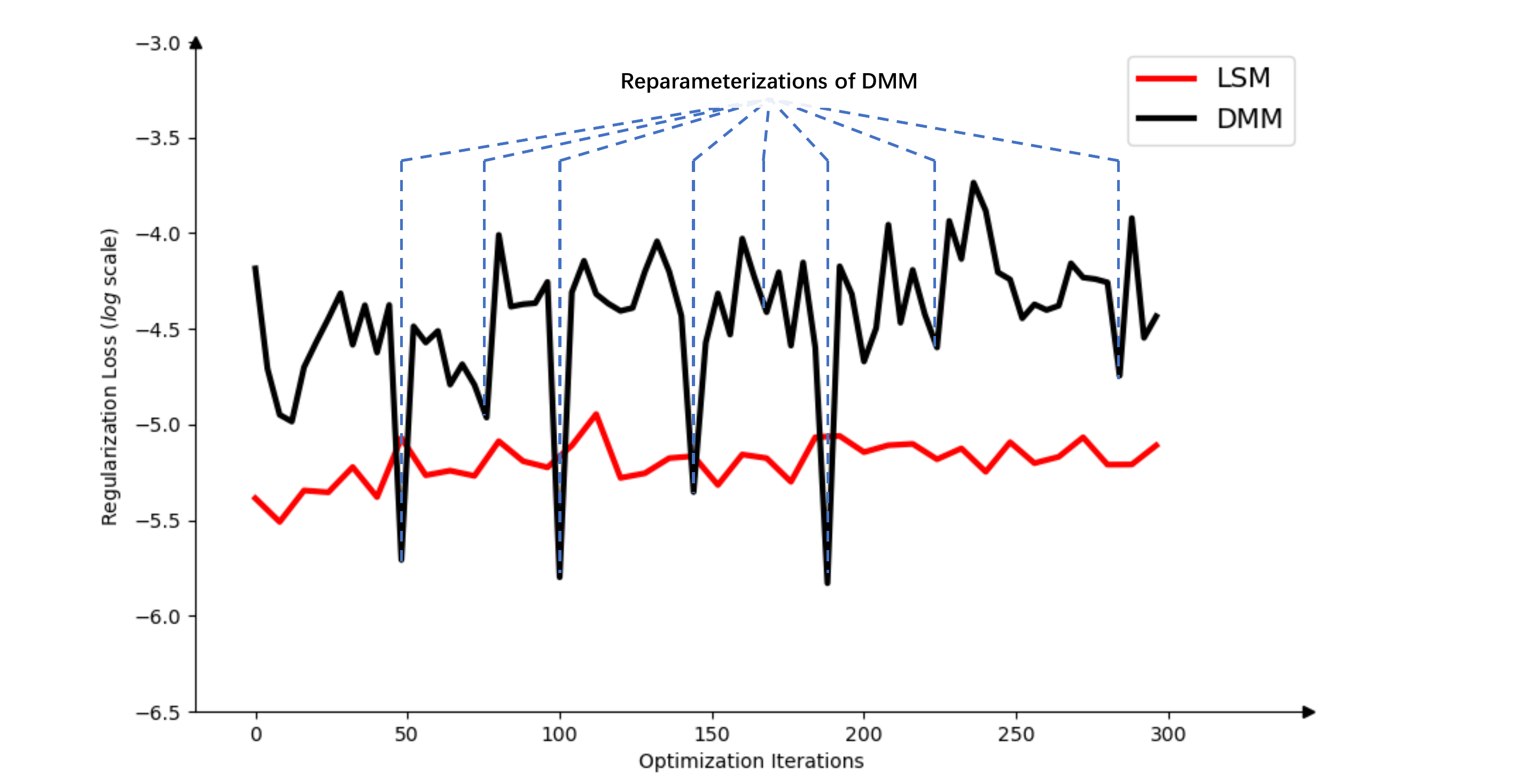}
    \end{center}
    \caption{
        \small The comparison of regularization losses of 'LSM+SU2' and 'DMM+SU2' settings. LSM deforms the computational mesh with better quality and less triggers the reparameterization.
    }
    \label{fig:reg_loss_compare}
\end{figure}

\begin{figure}[!htb]
    \begin{center}
        \includegraphics[width=0.9\linewidth]{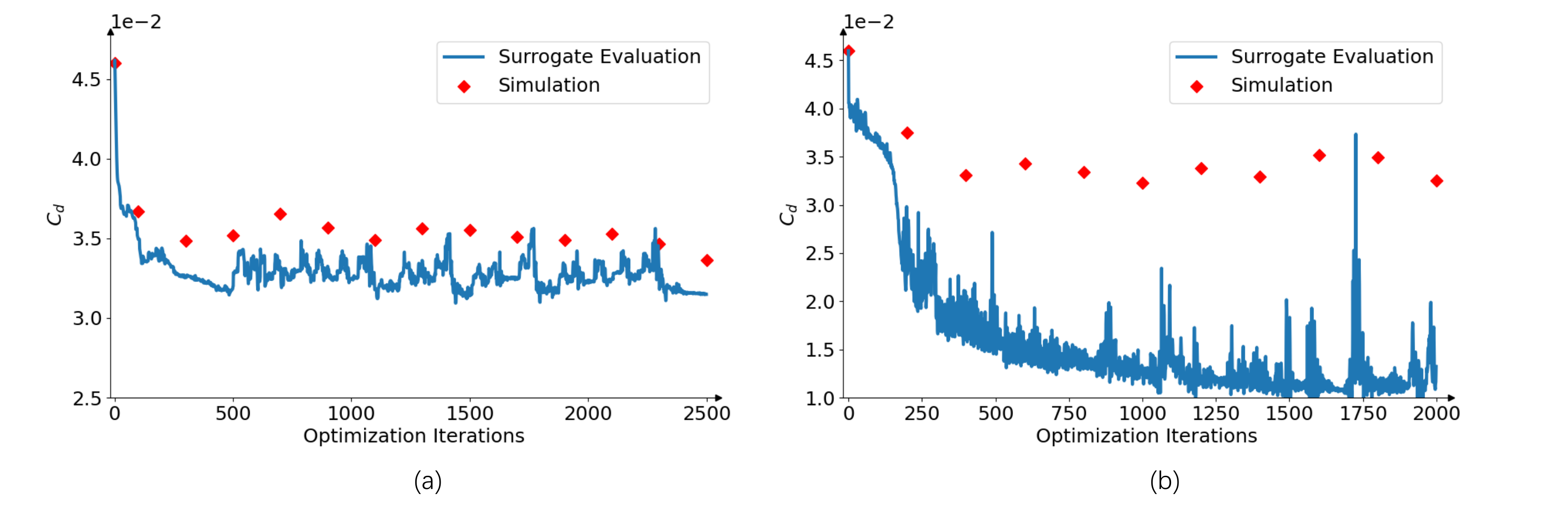}
    \end{center}
    \caption{
        \small The gap between GCNN's evaluations and simulation results becomes more significant when the airfoil is more deformed and falls outside the distribution of the training data, when it is combined with (a) LSM and (b) DMM. GCNN shows limited generalization ability when it only relies on a fixed training set.
    }
    \label{fig:gcnn_gap}
\end{figure}

The optimized NACA-0012 airfoils are shown in Fig.\ref{fig:final_optim_results}. The drag coefficients are significantly reduced in all four cases, indicating the effectiveness of the proposed parameterization methods with both evaluation models. The quantitative results in Tab.\ref{tab:result} demonstrate that the proposed automatic parameterization models perform comparably to previous handcrafted and semi-automatic parameterizations. However, the results differ between cases. Specifically, DMM generates larger deformations than LSM, which aligns with our discussion in Sec.\ref{sec:discussion_lsm_dmm} that the latent space of LSM parameterizes the distribution of the training dataset, while the optimal result obtained by \textit{DMM + SU2} does not follow the distribution of NACA profiles and the UIUC airfoil dataset. LSM provides simpler geometric constraints and better computational mesh deformation. As illustrated in Fig.\ref{fig:reg_loss_compare}, \textit{LSM + SU2} has a lower regularization loss value than \textit{DMM + SU2}, and no reparameterization is triggered during optimization.

In terms of evaluation models, SU2 consistently outperforms the GCNN model due to the limited generalization ability of GCNN. This limitation arises because the training set of GCNN includes fluid data only from existing airfoils. In this case, the optimized airfoils become out-of-distribution data, resulting in a large gap between the evaluation result and the real aerodynamic performance, as shown in Fig.\ref{fig:gcnn_gap}. This phenomenon makes it unsuitable for guiding the optimization process in the latter stage. However, SU2 takes considerably more computational time than GCNN. When both perform 100 iterations, using the GCNN model costs $9s$, while SU2 costs $2253s$ on the coarse template mesh \textit{TM-A}.

\section{Conclusion}

In this paper, we propose two deep learning models that fully automate shape parameterization for aerodynamic shape optimization. Both models eliminate the need for human intervention and directly manipulate high-dimensional mesh data. By learning continuous mappings of coordinate space, the models can deform both object surfaces and arbitrary types of computational meshes simultaneously. In the 2D airfoil optimization problem, our pipelines demonstrate comparable results with existing solutions based on handcrafted and semi-automatic parameterization methods. However, since the primary focus of this paper is on the parameterization model, other modules in the pipeline are not optimal. As an ongoing research project, we will continue to improve the overall performance by building better surrogate models and investigating meshes suitable for deformation. Meanwhile, we will continue to explore 3D object studies and apply deep geometric learning techniques to tackle more complex geometries.


\section*{Acknowledgement}
This work was supported in part by the Swiss National Science Foundation and the French “Programme d’Investissements D’avenir”: ANR-17-EURE-0005. Z. Wei is supported by the TSAE scholarship funded by Toulouse Graduate School in Aerospace Engineering. M. Bauerheim is supported by the French Direction Générale de l’Armement through the Agence de l’Innovation de Défense (AID) DECAP project.

\bibliography{string, sample}

\begin{thebibliography}{53}
\newcommand{\enquote}[1]{``#1''}
\providecommand{\natexlab}[1]{#1}
\providecommand{\url}[1]{\texttt{#1}}
\providecommand{\urlprefix}{URL }
\expandafter\ifx\csname urlstyle\endcsname\relax
  \providecommand{\doi}[1]{\discretionary{}{}{}https://doi.org/#1}\else
  \providecommand{\doi}[1]{\discretionary{}{}{}\urlstyle{rm}\url{https://doi.org/#1}}\fi

\bibitem[{Sederberg and Parry(1986)}]{Sederberg86}
Sederberg, T., and Parry, S., \enquote{{Free-Form Deformation of Solid
  Geometric Models},} \emph{ACM SIGGRAPH}, Vol.~20, No.~4, 1986.

\bibitem[{Lamousin and Jr.(1994)}]{Lamousin94}
Lamousin, H.~J., and Jr., W. N.~W., \enquote{{NURBS-based free-form
  deformations},} \emph{Computer Graphics and Applications}, Vol.~14, No.~6,
  1994, pp. 59--65.
\newblock \doi{10.1109/38.329096}.

\bibitem[{Kenway et~al.(2010)Kenway, Kennedy, and Martins}]{Kenway10}
Kenway, G., Kennedy, G., and Martins, J. R. R.~A., \enquote{A CAD-Free Approach
  to High-Fidelity Aerostructural Optimization,} \emph{13th AIAA/ISSMO
  Multidisciplinary Analysis Optimization Conference, Fort Worth, Texas, USA,
  September 13-15}, 2010.
\newblock \doi{10.2514/6.2010-9231}.

\bibitem[{Toal et~al.(2010)Toal, Bressloff, Keane, and Holden}]{Toal10}
Toal, D. J.~J., Bressloff, N.~W., Keane, A.~J., and Holden, C. M.~E.,
  \enquote{Geometric Filtration Using Proper Orthogonal Decomposition for
  Aerodynamic Design Optimization,} \emph{American Institute of Aeronautics and
  Astronautics Journal}, Vol.~48, No.~5, 2010, pp. 916--928.

\bibitem[{Kulfan(2008)}]{Kulfan08}
Kulfan, B.~M., \enquote{Universal Parametric Geometry Representation Method,}
  \emph{Journal of Aircraft}, Vol.~45, No.~1, 2008, pp. 142--158.
\newblock \doi{10.2514/1.29958}.

\bibitem[{{de Boer} et~al.(2007){de Boer}, {van der Schoot}, and
  Bijl}]{deBoer07}
{de Boer}, A., {van der Schoot}, M., and Bijl, H., \enquote{Mesh Deformation
  Based on Radial Basis Function Interpolation,} \emph{Computers and
  Structures}, Vol.~85, No.~11, 2007, pp. 784--795.

\bibitem[{Batina(1990)}]{Batina90}
Batina, J.~T., \enquote{Unsteady Euler airfoil solutions using unstructured
  dynamic meshes,} \emph{AIAA Journal}, Vol.~28, No.~8, 1990, pp. 1381--1388.
\newblock \doi{10.2514/3.25229}.

\bibitem[{Batina(1991)}]{Batina91}
Batina, J.~T., \enquote{Unsteady Euler algorithm with unstructured dynamic mesh
  for complex-aircraft aerodynamic analysis,} \emph{AIAA Journal}, Vol.~29,
  No.~3, 1991, pp. 327--333.
\newblock \doi{10.2514/3.10583}.

\bibitem[{Farhat et~al.(1998)Farhat, Degand, Koobus, and Lesoinne}]{Farhat98}
Farhat, C., Degand, C., Koobus, B., and Lesoinne, M., \emph{An improved method
  of spring analogy for dynamic unstructured fluid meshes}, 1998.
\newblock \doi{10.2514/6.1998-2070}.

\bibitem[{Luke et~al.(2012)Luke, Collins, and Blades}]{Luke12}
Luke, E., Collins, E., and Blades, E., \enquote{A fast mesh deformation method
  using explicit interpolation,} \emph{Journal of Computational Physics}, Vol.
  231, No.~2, 2012, pp. 586--601.
\newblock \doi{https://doi.org/10.1016/j.jcp.2011.09.021}.

\bibitem[{Bellman(1961)}]{Bellman61}
Bellman, R., \emph{Curse of Dimensionality}, Princeton university press, 1961.

\bibitem[{Robinson and Keane(2001)}]{Robinson01}
Robinson, G.~M., and Keane, A.~J., \enquote{Concise Orthogonal Representation
  of Supercritical Airfoils,} \emph{Journal of Aircraft}, Vol.~38, No.~3, 2001,
  pp. 580--583.
\newblock \doi{10.2514/2.2803}.

\bibitem[{Poole et~al.(2015{\natexlab{a}})Poole, Allen, and Rendall}]{Poole15}
Poole, D.~J., Allen, C.~B., and Rendall, T. C.~S., \enquote{Metric-Based
  Mathematical Derivation of Efficient Airfoil Design Variables,}
  \emph{American Institute of Aeronautics and Astronautics Journal}, Vol.~53,
  No.~5, 2015{\natexlab{a}}, pp. 1349--1361.
\newblock \doi{10.2514/1.J053427}.

\bibitem[{Masters et~al.(2017)Masters, Taylor, Rendall, Allen, and
  Poole}]{Masters17}
Masters, D.~A., Taylor, N.~J., Rendall, T. C.~S., Allen, C.~B., and Poole,
  D.~J., \enquote{Geometric Comparison of Aerofoil Shape Parameterization
  Methods,} \emph{American Institute of Aeronautics and Astronautics Journal},
  Vol.~55, No.~5, 2017, pp. 1575--1589.
\newblock \doi{10.2514/1.J054943}.

\bibitem[{Li et~al.(2019{\natexlab{a}})Li, Bouhlel, and Martins}]{Li19c}
Li, J., Bouhlel, M.~A., and Martins, J. R. R.~A., \enquote{Data-Based Approach
  for Fast Airfoil Analysis and Optimization,} \emph{American Institute of
  Aeronautics and Astronautics Journal}, Vol.~57, No.~2, 2019{\natexlab{a}},
  pp. 581--596.
\newblock \doi{10.2514/1.J057129}.

\bibitem[{Kedward et~al.(2020)Kedward, Allen, and Rendall}]{Kedward20}
Kedward, L., Allen, C.~B., and Rendall, T., \enquote{Towards Generic Modal
  Design Variables for Aerodynamic Shape Optimisation,} \emph{AIAA Scitech
  Forum, Orlando, FL, USA, January 6-10}, 2020.
\newblock \doi{10.2514/6.2020-0543}.

\bibitem[{Constantine et~al.(2014)Constantine, Dow, and Wang}]{Constantine14}
Constantine, P.~G., Dow, E., and Wang, Q., \enquote{{Active Subspace Methods in
  Theory and Practice: Applications to Kriging Surfaces},} \emph{SIAM Journal
  on Scientific Computing}, Vol.~36, No.~4, 2014, pp. A1500--A1524.
\newblock \doi{10.1137/130916138}.

\bibitem[{Li et~al.(2019{\natexlab{b}})Li, Cai, and Qu}]{Li19}
Li, J., Cai, J., and Qu, K., \enquote{{Surrogate-based aerodynamic shape
  optimization with the active subspace method},} \emph{{Structural and
  Multidisciplinary Optimization}}, Vol.~59, No.~2, 2019{\natexlab{b}}, pp.
  403--419.
\newblock \doi{10.1007/s00158-018-2073-5}.

\bibitem[{Lukaczyk et~al.(2014)Lukaczyk, Constantine, Palacios, and
  Alonso}]{Lukaczyk14}
Lukaczyk, T.~W., Constantine, P., Palacios, F., and Alonso, J.~J.,
  \enquote{Active Subspaces for Shape Optimization,} \emph{10th AIAA
  Multidisciplinary Design Optimization Conference}, 2014.
\newblock \doi{10.2514/6.2014-1171}.

\bibitem[{Namura et~al.(2017)Namura, Shimoyama, and Obayashi}]{Namura17}
Namura, N., Shimoyama, K., and Obayashi, S., \enquote{Kriging surrogate model
  with coordinate transformation based on likelihood and gradient,}
  \emph{Journal of Global Optimization}, Vol.~68, No.~3, 2017, pp. 827--849.
\newblock \doi{10.1007/s10898-017-0516-y}.

\bibitem[{Grey and Constantine(2018)}]{Grey18}
Grey, Z.~J., and Constantine, P.~G., \enquote{Active Subspaces of Airfoil Shape
  Parameterizations,} \emph{AIAA Journal}, Vol.~56, No.~5, 2018, pp.
  2003--2017.
\newblock \doi{10.2514/1.J056054}.

\bibitem[{Bauerheim et~al.(2016)Bauerheim, Ndiaye, Constantine, Moreau, and
  Nicoud}]{Bauerheim16}
Bauerheim, M., Ndiaye, A., Constantine, P., Moreau, S., and Nicoud, F.,
  \enquote{Symmetry breaking of azimuthal thermoacoustic modes: the UQ
  perspective,} \emph{Journal of Fluid Mechanics}, Vol. 789, 2016, p.
  534–566.
\newblock \doi{10.1017/jfm.2015.730}.

\bibitem[{Magri et~al.(2016)Magri, Bauerheim, Nicoud, and Juniper}]{Magri16}
Magri, L., Bauerheim, M., Nicoud, F., and Juniper, M.~P., \enquote{Stability
  analysis of thermo-acoustic nonlinear eigenproblems in annular combustors.
  Part II. Uncertainty quantification,} \emph{Journal of Computational
  Physics}, Vol. 325, 2016, pp. 411--421.

\bibitem[{Viswanath et~al.(2011)Viswanath, Forrester, and Keane}]{Viswanath11}
Viswanath, A., Forrester, A. I.~J., and Keane, A.~J., \enquote{Dimension
  Reduction for Aerodynamic Design Optimization,} \emph{American Institute of
  Aeronautics and Astronautics Journal}, Vol.~49, No.~6, 2011, pp. 1256--1266.

\bibitem[{Li et~al.(2020)Li, Zhang, Martins, and Shu}]{li20}
Li, J., Zhang, M., Martins, J. R. R.~A., and Shu, C., \enquote{Efficient
  Aerodynamic Shape Optimization with Deep-Learning-Based Geometric Filtering,}
  \emph{American Institute of Aeronautics and Astronautics Journal}, Vol.~58,
  No.~10, 2020, pp. 4243--4259.
\newblock \doi{10.2514/1.J059254}.

\bibitem[{Li and Zhang(2021)}]{Li21}
Li, J., and Zhang, M., \enquote{On deep-learning-based geometric filtering in
  aerodynamic shape optimization,} \emph{Aerospace Science and Technology},
  Vol. 112, 2021, p. 106603.

\bibitem[{Li et~al.(2022)Li, Zhang, and Chen}]{Li22b}
Li, R., Zhang, Y., and Chen, H., \enquote{Physically Interpretable Feature
  Learning of Supercritical Airfoils Based on Variational Autoencoders,}
  \emph{AIAA Journal}, Vol.~60, No.~11, 2022, pp. 6168--6182.
\newblock \doi{10.2514/1.J061673}.

\bibitem[{Glaws et~al.(2022)Glaws, King, Vijayakumar, and Ananthan}]{Glaws22}
Glaws, A., King, R.~N., Vijayakumar, G., and Ananthan, S., \enquote{Invertible
  Neural Networks for Airfoil Design,} \emph{AIAA Journal}, Vol.~60, No.~5,
  2022, pp. 3035--3047.
\newblock \doi{10.2514/1.J060866}.

\bibitem[{Park et~al.(2019)Park, Florence, Straub, Newcombe, and
  Lovegrove}]{Park19c}
Park, J.~J., Florence, P., Straub, J., Newcombe, R.~A., and Lovegrove, S.,
  \enquote{{DeepSdf: Learning Continuous Signed Distance Functions for Shape
  Representation},} \emph{Conference on Computer Vision and Pattern
  Recognition}, 2019.

\bibitem[{Tan and Mayrovouniotis(1995)}]{Tan95}
Tan, S., and Mayrovouniotis, M.~L., \enquote{{Reducing data dimensionality
  through optimizing neural network inputs},} \emph{AIChE Journal}, Vol.~41,
  No.~6, 1995, pp. 1471--1480.

\bibitem[{Barrow et~al.(1977)Barrow, Tenenbaum, Bolles, and Wolf}]{Barrow77}
Barrow, H.~G., Tenenbaum, J.~M., Bolles, R.~C., and Wolf, H.~C.,
  \enquote{Parametric Correspondence and Chamfer Matching: Two New Techniques
  for Image Matching,} \emph{International Joint Conference on Artificial
  Intelligence}, 1977.

\bibitem[{Knupp(2007)}]{knupp07}
Knupp, P., \enquote{Remarks on Mesh Quality,} Tech. rep., Sandia National
  Lab.(SNL-NM), Albuquerque, NM (United States), 2007.

\bibitem[{Shen and Zhou(2021)}]{shen21}
Shen, Y., and Zhou, B., \enquote{Closed-Form Factorization of Latent Semantics
  in GANs,} \emph{Conference on Computer Vision and Pattern Recognition}, 2021.
\newblock \doi{10.1109/cvpr46437.2021.00158}.

\bibitem[{Wei et~al.(2023)Wei, Guillard, Fua, Chapin, and Bauerheim}]{wei23}
Wei, Z., Guillard, B., Fua, P., Chapin, V., and Bauerheim, M., \enquote{Latent
  Representation of CFD Meshes and Application to 2D Airfoil Aerodynamics,}
  \emph{AIAA Journal}, 2023.

\bibitem[{Paszke et~al.(2019)Paszke, Gross, Massa, Lerer, Bradbury, Chanan,
  Killeen, Lin, Gimelshein, Antiga et~al.}]{paszke19}
Paszke, A., Gross, S., Massa, F., Lerer, A., Bradbury, J., Chanan, G., Killeen,
  T., Lin, Z., Gimelshein, N., Antiga, L., et~al., \enquote{PyTorch: An
  Imperative Style, High-Performance Deep Learning Library,} \emph{Advances in
  Neural Information Processing Systems}, Vol.~32, 2019, pp. 8024--8035.

\bibitem[{Fukushima(1975)}]{fukushima75}
Fukushima, K., \enquote{Cognitron: A self-organizing multilayered neural
  network,} \emph{Biological Cybernetics}, Vol.~20, No. 3-4, 1975, pp.
  121--136.

\bibitem[{Sitzmann et~al.(2020)Sitzmann, Martel, Bergman, Lindell, and
  Wetzstein}]{sitzmann20}
Sitzmann, V., Martel, J. N.~P., Bergman, A.~W., Lindell, D.~B., and Wetzstein,
  G., \enquote{Implicit Neural Representations with Periodic Activation
  Functions,} \emph{Advances in Neural Information Processing Systems},
  Vol.~33, 2020, pp. 7462--7473.

\bibitem[{Selig(1996)}]{Selig96a}
Selig, M., \emph{UIUC airfoil data site}, Department of Aeronautical and
  Astronautical Engineering, University of Illinois at Urbana-Champaign, 1996.

\bibitem[{Kingma and Ba(2015)}]{Kingma15}
Kingma, D.~P., and Ba, J., \enquote{{Adam: {A} Method for Stochastic
  Optimisation},} \emph{International Conference on Learning Representations},
  2015.

\bibitem[{Baqu{\'{e}} et~al.(2018)Baqu{\'{e}}, Remelli, Fleuret, and
  Fua}]{Baque18}
Baqu{\'{e}}, P., Remelli, E., Fleuret, F., and Fua, P., \enquote{{Geodesic
  Convolutional Shape Optimization},} \emph{International Conference on Machine
  Learning}, 2018.

\bibitem[{Economon et~al.(2016)Economon, Palacios, Copeland, Lukaczyk, and
  Alonso}]{Economon16}
Economon, T.~D., Palacios, F., Copeland, S.~R., Lukaczyk, T.~W., and Alonso,
  J.~J., \enquote{SU2: An Open-Source Suite for Multiphysics Simulation and
  Design,} \emph{American Institute of Aeronautics and Astronautics Journal},
  2016.
\newblock \doi{10.2514/1.J053813}.

\bibitem[{Ioffe and Szegedy(2015)}]{Sergey15}
Ioffe, S., and Szegedy, C., \enquote{Batch Normalization: Accelerating Deep
  Network Training by Reducing Internal Covariate Shift,} \emph{Proceedings of
  the 32nd International Conference on Machine Learning}, Vol.~37, PMLR, 2015,
  pp. 448--456.

\bibitem[{Clevert et~al.(2016)Clevert, Unterthiner, and Hochreiter}]{clevert16}
Clevert, D.-A., Unterthiner, T., and Hochreiter, S., \enquote{Fast and Accurate
  Deep Network Learning by Exponential Linear Units (ELUs),}
  \emph{International Conference on Learning Representations}, 2016.

\bibitem[{LeDoux et~al.(2015)LeDoux, Vassberg, Young, Fugal, Kamenetskiy,
  Huffman, Melvin, and Smith}]{LeDoux15}
LeDoux, S.~T., Vassberg, J.~C., Young, D.~P., Fugal, S., Kamenetskiy, D.,
  Huffman, W.~P., Melvin, R.~G., and Smith, M.~F., \enquote{Study Based on the
  AIAA Aerodynamic Design Optimization Discussion Group Test Cases,} \emph{AIAA
  Journal}, Vol.~53, No.~7, 2015, pp. 1910--1935.
\newblock \doi{10.2514/1.J053535}.

\bibitem[{He et~al.(2019)He, Li, Mader, Yildirim, and Martins}]{he19}
He, X., Li, J., Mader, C.~A., Yildirim, A., and Martins, J.~R., \enquote{Robust
  aerodynamic shape optimization—From a circle to an airfoil,}
  \emph{Aerospace Science and Technology}, Vol.~87, 2019, pp. 48--61.
\newblock \doi{10.1016/j.ast.2019.01.051}.

\bibitem[{Masters et~al.(16)Masters, Poole, Taylor, Rendall, and
  Allen}]{Masters16}
Masters, D.~A., Poole, D.~J., Taylor, N.~J., Rendall, T., and Allen, C.~B.,
  \emph{Impact of Shape Parameterisation on Aerodynamic Optimisation of
  Benchmark Problem}, 16.
\newblock \doi{10.2514/6.2016-1544}.

\bibitem[{Bisson and Nadarajah(2015)}]{Bisson15}
Bisson, F., and Nadarajah, S., \emph{Adjoint-Based Aerodynamic Optimization of
  Benchmark Problems}, 2015.
\newblock \doi{10.2514/6.2015-1948}.

\bibitem[{Carrier et~al.(2014)Carrier, Destarac, Dumont, Meheut, Din, Peter,
  Khelil, Brezillon, and Pestana}]{Carrier14}
Carrier, G., Destarac, D., Dumont, A., Meheut, M., Din, I. S.~E., Peter, J.,
  Khelil, S.~B., Brezillon, J., and Pestana, M., \emph{Gradient-Based
  Aerodynamic Optimization with the elsA Software}, 2014.
\newblock \doi{10.2514/6.2014-0568}.

\bibitem[{Lee et~al.(2015)Lee, Koo, Telidetzki, Buckley, Gagnon, and
  Zingg}]{Lee15a}
Lee, C., Koo, D., Telidetzki, K., Buckley, H., Gagnon, H., and Zingg, D.,
  \enquote{{Aerodynamic Shape Optimization of Benchmark Problems Using
  Jetstream},} \emph{53rd AIAA Aerospace Sciences Meeting}, 2015.

\bibitem[{Zhang et~al.(2016)Zhang, Han, Shi, and Song}]{Zhang16}
Zhang, Y., Han, Z.-H., Shi, L., and Song, W.-P., \emph{Multi-round
  Surrogate-based Optimization for Benchmark Aerodynamic Design Problems},
  2016.
\newblock \doi{10.2514/6.2016-1545}.

\bibitem[{Poole et~al.(2015{\natexlab{b}})Poole, Allen, and Rendall}]{Poole15a}
Poole, D.~J., Allen, C.~B., and Rendall, T., \emph{Control Point-Based
  Aerodynamic Shape Optimization Applied to AIAA ADODG Test Cases},
  2015{\natexlab{b}}.
\newblock \doi{10.2514/6.2015-1947}.

\bibitem[{Gariepy et~al.(2015)Gariepy, Trepanier, Petro, Malouin, Audet,
  LeDigabel, and Tribes}]{Gariepy15}
Gariepy, M., Trepanier, J.-Y., Petro, E., Malouin, B., Audet, C., LeDigabel,
  S., and Tribes, C., \emph{Direct Search Airfoil Optimization Using Far-Field
  Drag Decomposition Results}, 2015.
\newblock \doi{10.2514/6.2015-1720}.

\bibitem[{Fabiano and Mavriplis(2016)}]{Fabiano16}
Fabiano, E., and Mavriplis, D.~J., \emph{Adjoint-Based Aerodynamic Design On
  Unstructured Meshes}, 2016.
\newblock \doi{10.2514/6.2016-1295}.

\end{thebibliography}

\end{document}